\documentclass[journal]{IEEEtran}
\usepackage{booktabs}
\usepackage{multirow}
\usepackage[table,xcdraw]{xcolor}

\usepackage[pdftex]{graphicx}
\usepackage{subcaption} 
\ifCLASSINFOpdf
\else
\fi
\usepackage{tikz}
\usepackage{circuitikz}
\usepackage{svg}

\usepackage{amsmath}
\usepackage{amsfonts}
\usepackage{xcolor}
\newcommand{\rev}[1]{\textcolor{black}{#1}}
\newenvironment{revblock}{\color{black}}{}

\usepackage{mdframed}
\usepackage{ntheorem}
\theoremstyle{nonumberplain}
\newmdtheoremenv[%
  linecolor=blue,
  linewidth=2pt,
  rightline=false,
  leftline=false]{figrev}{}

\usepackage[export]{adjustbox}

\begin{document}

%
\title{Locally Linear Continual Learning for Time Series based on VC-Theoretical Generalization Bounds}
%
%
%
%

%

\author{Yan V. G. Ferreira, 
        Igor B. Lima,
        Pedro H. G. Mapa S.,
        Felipe V. Campos,
        and Antonio P. Braga
\thanks{Yan V. G. Ferreira, Igor B. Lima and Antonio P. Braga are with the Department of Electronic Engineering, Universidade Federal de Minas Gerais, UFMG, Belo Horizonte 31270-901, Brazil (e-mail: eng-yanvictor@ufmg.br, igorbraga@ufmg.br, apbraga@ufmg.br).}
\thanks{Pedro H. G. Mapa S. is with the Department of Computing Science, University of Alberta, Edmonton, Alberta, Canada (e-mail: gomesmap@ualberta.ca)}
\thanks{Felipe V. Campos and Antonio P. Braga are with the Graduate Program in Electrical Engineering, Universidade Federal de Minas Gerais, UFMG, Belo Horizonte 31270-901, Brazil (e-mail: felipevellosoc@ufmg.br)}}

\maketitle

\begin{abstract}

Most machine learning methods assume fixed probability distributions, limiting their applicability in nonstationary real-world scenarios. While continual learning methods address this issue, current approaches often rely on black-box models or require extensive user intervention for interpretability. We propose SyMPLER (Systems Modeling through Piecewise Linear Evolving Regression), an explainable model for time series forecasting in nonstationary environments based on dynamic piecewise-linear approximations. Unlike other locally linear models, SyMPLER uses generalization bounds from Statistical Learning Theory to automatically determine when to add new local models based on prediction errors, eliminating the need for explicit clustering of the data. Experiments show that SyMPLER can achieve comparable performance to both black-box and existing explainable models while maintaining a human-interpretable structure that reveals insights about the system's behavior. In this sense, our approach conciliates accuracy and interpretability, offering a transparent and adaptive solution for forecasting nonstationary time series.

%
%
%
%

\end{abstract}

\begingroup
\renewcommand\thefootnote{}
\footnotetext{© 2026 IEEE. Personal use is permitted.
The final version of record is available at https://doi.org/10.1109/TPAMI.2026.3672726}
\endgroup


%

%
\section{Introduction}


\IEEEPARstart{L}{earning} in dynamic environments requires models not only to capture the system's behavior at a given moment, but also to adapt when conditions change. This is usual in real-world scenarios, where factors such as weather changes, wear and tear in mechanical systems, or variations in user behavior can alter system dynamics. Continual Learning (CL) is an active research area in machine learning that addresses this challenge by enabling models to estimate functions that evolve over time or whose domains change dynamically \cite{defying-forgetting}. When the relationship between inputs and outputs changes, the problem is known as concept drift, often caused by machinery aging or shifts in system objectives \cite{carla}. Conversely, if the input distribution shifts while the functional relationship remains the same, it is referred to as covariate drift \cite{carla} or domain-incremental learning \cite{three-types}. This typically occurs when input values enter previously unseen ranges, such as during operational changes in an industrial plant or when an autonomous system explores unfamiliar environments. In this study, we place particular emphasis on the domain-incremental setting.

Since the generator function of most real-world problems is non-linear, it is difficult to extrapolate the knowledge obtained at a region of the domain to other regions. One way to adapt to new domains is to update the model’s parameters to minimize the error on incoming data. However, this typically comes at the cost of losing performance on previously learned regions, a phenomenon known as catastrophic forgetting \cite{catastrophic-forgetting}. Thus, a continual learning model must balance adaptability (plasticity) with retention of prior knowledge (stability). This trade-off, known in the machine learning literature as the stability-plasticity dilemma \cite{defying-forgetting}, is a fundamental challenge in online learning.


Many approaches to deal with such adversities have been proposed, mainly focused on modifications on the structure or on the training procedure of artificial neural networks (ANNs) \cite{defying-forgetting, first-survey}. However, since ANNs usually lack interpretability, these approaches yield black-box models, unsuitable for high-stakes scenarios, such as in industry and medical applications. Interpretable alternatives exist, such as evolving fuzzy approaches \cite{survey-fuzzy}, which produce human-interpretable rules for local regions of the input. Although these methods produce accurate and explainable results, they usually depend on the explicit clustering of the input data, which may have user-defined thresholds for adding new local rules, and have hyperparameters related to the shape of the fuzzy membership functions or the update of the rules' parameters.

In the present work, we propose SyMPLER, a bottom-up constructive approach for domain-incremental time series forecasting, aiming at an explainable global approximation through localized linear functions. Our main contributions are:

\begin{enumerate}
    \item Explainability through a piecewise-linear structure.
    \item Clustering-free model evolution based on performance.
    \item Theoretical guarantees on locality and generalization based on Statistical Learning Theory.
    \item Competitive performance with black-box models.
\end{enumerate}

The work is organized as follows. Section \ref{sec:related} explores the continual learning problem and the main existing approaches to solving it. Section \ref{sec:methods} presents SyMPLER. The method is applied both in a simulated environment and in a real problem in Section \ref{sec:exps}, with some cases of interest further explored in Section \ref{sec:add-exp}. Finally, the main findings of this work are summarized in Section \ref{sec:conclusion}.
\section{Related Works}
\label{sec:related}

%

%
%
Several methods have been proposed to handle learning under covariate and concept drifts. However, as noted by He and Sick \cite{clear} and Besnard and Ragot \cite{first-survey}, most approaches focus on classification tasks, with a detailed taxonomy provided by Delange et al. \cite{defying-forgetting}. Continual learning for regression, especially in time series forecasting, remains less explored. 

%
Some of the methods developed for classification have been extended for regression, such as Elastic Weight Consolidation (EWC) \cite{ewc} training algorithm for artificial neural networks (ANNs). This algorithm was proposed to overcome catastrophic forgetting in continual learning by penalizing the change of weights important to previous tasks. By doing this, EWC favors adapting different groups of weights for different tasks, allowing the network to learn new patterns without forgetting previous ones. Zhou et al. \cite{ewc-based} applied an EWC-based sliding window fine tuning (EWC-SWFT) method to train a neural network for online prediction of energy loads in a chinese building, obtaining better performance than other continual learning algorithms for ANNs. However, EWC-based methods require the user to explicitly define the tasks the model should retain. In streaming data scenarios, such as time series forecasting, task boundaries are often unclear, and it is generally preferable for models to detect changes and adapt autonomously.

Although these methods produce state-of-the-art models with great predictive power along with flexibility and little forgetfulness for continual learning, they lack interpretability and explainability, since they rely on black-box neural networks as base models. Another approach for the continual learning problem is based on piecewise-linear approximations of the non-linear generator function, which are fully interpretable by construction. Fuzzy Takagi-Sugeno models are a very popular class of these models both in control and machine learning literature. They use linear rules for fuzzily defined local regions of the input space, allowing modeling based, for example, on linguistic descriptions of the system. This framework has long been adapted for automatic continual learning purposes, as reviewed in detail by Škrjanc et al. \cite{survey-fuzzy}. Evolving fuzzy (or neuro-fuzzy) models define local regions adaptively, usually based on online clustering techniques for the streaming data, and adjust the parameters of its linear rules with recursive least-squares estimates. (See, for instance, eTS (Evolving Takagi-Sugeno) models \cite{ets} and its variants, such as simpl\_eTS \cite{simplets}.) Most of these methods, however, still depend on user-defined parameters -- especially thresholds for the creation of new rules --, which can be hard to adjust for each new problem. Furthermore, explicit clustering of the input data adds complexity and computational overhead to the methods, in addition to not ensuring the locality of the rules depending on the number of centers added.

\begin{revblock}
    
Motivated by the interpretability of local linear models, SyMPLER adaptively defines approximation points and local weights without relying on explicit clustering techniques. The approach is related to Ronco and Gawthrop’s Incremental Controller Networks (ICN) \cite{icn}, which evolve a gated network of linear models from streaming data, using the current prediction error as an indicator of unmodeled regions. However, ICN was originally designed for control applications and assumes that the plant can be halted to actively collect data when a new operating region is detected, an assumption often unrealistic in online learning settings. In addition, ICN relies on a user-defined error threshold for model addition and provides no theoretical guarantees on the performance of individual linear models. In contrast, SyMPLER enables fully automatic model addition through an user-independent criterion that compares the network prediction error to that of a naïve baseline. Combined with incremental learning from streaming data, this allows model growth without user intervention or system interruption. Moreover, grounded in Statistical Learning Theory, SyMPLER defines a theoretically optimal novelty buffer that guarantees generalization of local models with minimal training data, ensuring their local validity.
%

\end{revblock}

\section{SyMPLER}
\label{sec:methods}

\begin{figure}[t] 
    \centering
    \begin{subfigure}{\linewidth}
        \centering
        \includegraphics[width=0.635\linewidth]{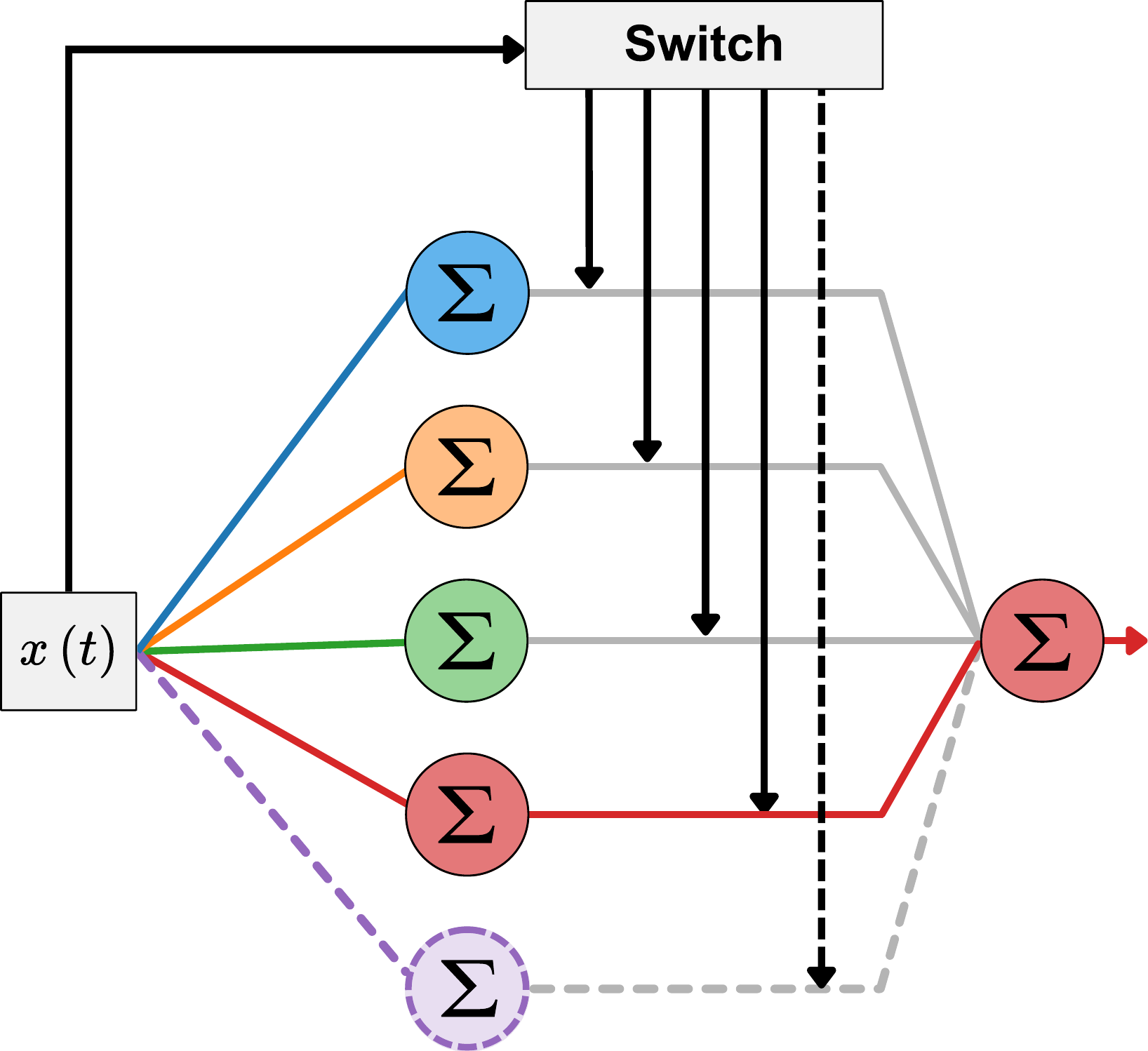}
        \label{fig:sympler}
    \end{subfigure}

    \begin{subfigure}{\linewidth}
        \centering
        \includegraphics[width=0.8\linewidth]{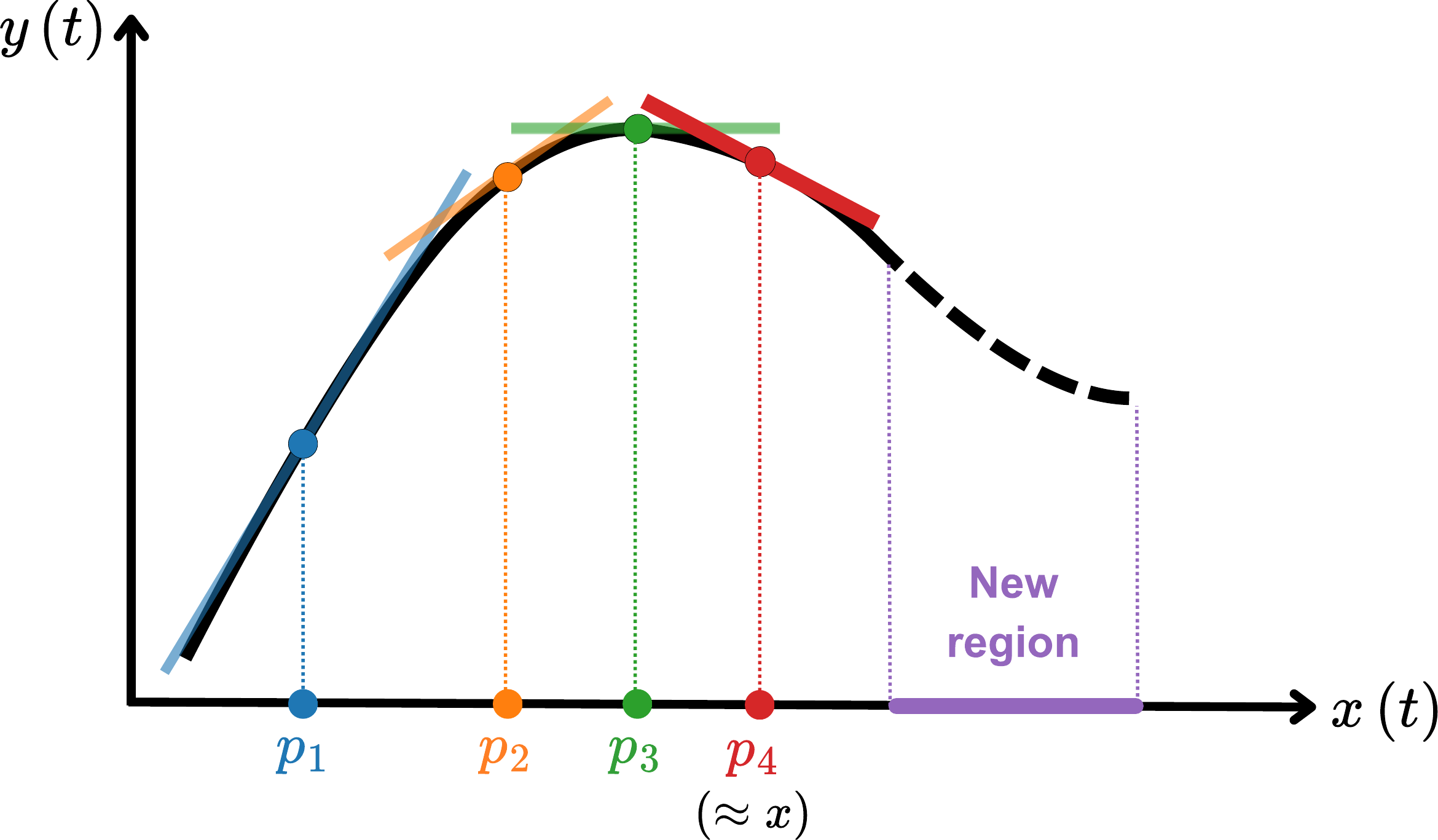}
        \label{fig:sympler-graph}
    \end{subfigure}
    
    \caption{SyMPLER switches between existing local models depending on the value of the input $\mathbf{x}(t)$. Each local model approximates the desired function linearly around its approximation point $\mathbf{x}(t) = \mathbf{p}_i$. As the domain of the problem grows, more local models are added. In this example, SyMPLER switches to the model colored in red as $\mathbf{x}(t)$ approaches $\mathbf{p}_4$ and adds a new model for the new region of the input.}
    \label{fig:sympler-full}
\end{figure}

SyMPLER can be seen as a gated neural network with one hidden layer of dynamic size and linear activation functions, as illustrated in Figure \ref{fig:sympler-full}. The $i$-th neuron of the network corresponds to a linear model in input space which approximates the desired function in the neighborhood of the approximation point $\mathbf{x}(t)=\mathbf{p}_i$. To model the input-output dependency completely, SyMPLER switches between local models depending on the value of the input, allowing it to approximate non-linear functions piecewise-linearly. When new data arrive from a region of the input never seen before, new local models are added to model it, growing the structure of the network. Conversely, when data arrive from previously seen regions, the model only has to switch to the local model trained in that region. Thus, SyMPLER suffers very little from the stability-plasticity dilemma, maintaining previous knowledge while still learning new regions of the problem. In addition to this, since each local model is linear in the input space, SyMPLER is fully interpretable and explainable. In this sense, the present section describes the theoretical foundations of SyMPLER and its algorithm for creating new local models and switching between existing ones.


\subsection{Piecewise-Linear Approximation}
\label{subsec:piecewise}

Taylor's theorem states that  any function $f: \mathbb{R} \rightarrow \mathbb{R}$ at least $k$ times differentiable at the point $x=x_0$ can be approximated by the Taylor series truncated at the $k$-th term. That is,

\begin{equation}
    f(x) = \sum^k_{i=0}\frac{f^{(i)}(x_0)}{i!}(x-x_0)^i + R_k(x),
\label{eq:taylor-univariate}
\end{equation}
where $R_k(x)$ -- the approximation error -- tends to zero as $x$ approaches $x_0$. If the $k$-th derivative, $f^{(k)}$, is continuous on the closed interval between $x_0$ and $x$, there exists some real number $c$ in this interval such that

\begin{equation}
    R_k(x) = \frac{f^{(k+1)}(c)}{(k+1)!}(x-x_0)^{k+1},
    \label{eq:lagrange-error}
\end{equation}
which is called the Lagrange form of the approximation error.

For interpretability purposes, it is particularly interesting that $f$ can be approximated by a line around a point $x=x_0$ if it is differentiable at that point. From \eqref{eq:lagrange-error}, it follows that the approximation error is proportional to the square of the difference between $x$ and $x_0$. Then, to approximate $f$ with an arbitrarily small error outside the neighborhood of $x_0$ one can choose other points on the domain, $\{x_1, x_2, \dots, x_n\}$, and use the first order Taylor polynomials at those points. In this sense, a non-linear function can be arbitrarily approximated using several local linear functions, each one corresponding to a region of the domain.

In the multivariate case, for a function $f: \mathbb{R}^n \rightarrow \mathbb{R}$, the first-order Taylor approximation around a point $\mathbf{x_0} \in \mathbb{R}^n$ is given in terms of its gradient $\mathbf{\nabla f}$ at that point, which is equivalent to approximating the function by its tangent hyperplane. In this sense, the function can be expressed as

\begin{equation}
    f(\mathbf{x}) = \mathbf{\nabla f}(\mathbf{x_0})(\mathbf{x}-\mathbf{x_0}) + o(||\mathbf{x}-\mathbf{x_0}||^2),
\label{eq:taylor-multivariate}
\end{equation}
where $o(||\mathbf{x}-\mathbf{x_0}||^2)$ denotes a term that decreases more slowly than $||\mathbf{x}-\mathbf{x_0}||^2$ as $\mathbf{x}$ approaches $\mathbf{x_0}$. Therefore, the approximation error also depends on the distance between a test point and the approximation point in this case.

This fact has been extensively explored for model linearization, specially in control \cite{narmax, billings} and machine learning \cite{split-regularized-regression, locally-linear-ensemble, combination-of-ensembles, ets, simplets, panfis, icn} literature. A committee of several local ``specialist" models is trained: when the input data is in the region they model, its output dominates the output of the committee. SyMPLER follows the same strategy, but creating new local models as the domain of the problem changes. 

\subsection{Adding New Models}
\label{subsec:grow}

A common challenge when building locally linear models is selecting appropriate approximation points. In fuzzy systems, for instance, this is usually done considering expert knowledge about the operating regimes of the system or via explicit clustering of the input data. However, expert input can be unfeasible in complex and high-dimensional settings, while clustering techniques typically requires an user-defined threshold for adding new rules \cite{survey-fuzzy}.

SyMPLER addresses these limitations by autonomously selecting approximation points and adding local models without external inputs or prior knowledge about the system. This is achieved comparing the model's error to that of a baseline model. In time series problems, a natural baseline is the naïve predictor, which predicts the output at time $t+1$ to be equal the output at time $t$. Since it is always desirable to overcome the performance of those baseline models, approximation points should be added in the regions where the error of the current network is greater than the baseline's. Our approach is, then, to compare the error of both models through time and store the data where the current model performs worse than the baseline in a retraining buffer. Once enough data -- according to a criterion that will be further explained -- are collected, a new local linear model is trained as an specialist for that region of the input. Although several possible baselines could be considered, we argue that the delayed predictor is the most appropriate for time-series predictions, specially short-term ones. In particular, if the time series being predicted varies slowly through time (in relation to its sampling period), using the most recent information about the output usually provides good forecasting results without any training.

\begin{revblock}

However, comparing the models’ errors on a sample-by-sample basis makes the update criterion highly sensitive to noise. Consider, for instance, a stream of $N$ samples in which the network performs worse than the baseline in $N-1$ cases but, by chance, achieves a lower error in a single sample. Despite this isolated improvement, the model should still be updated.

To mitigate this sensitivity, we propose comparing the models using the cumulative average of their errors rather than instantaneous values. Specifically, we start tracking cumulative averages from the first sample $(x_0, y_0)$ for which the network’s error exceeds that of the baseline, and this sample initializes a training buffer for a potential new model. As new samples $(x_i, y_i)$ arrive, they are added to the buffer as long as the average network error, computed over the buffered samples, remains greater than the baseline’s average error over the same interval. Assuming that data are sampled at a rate much faster than changes in the system’s operating point, this procedure can be interpreted as a comparison of the expected errors of the two models at that operating condition.

A new model is trained only if the buffer contains a sufficient number of samples to fit a linear model without overfitting. If, before the buffer reaches this minimum size, the cumulative average error of the baseline becomes lower than that of the network, the buffered samples are discarded, as they do not provide enough information to justify training a new model. As we will show later, this limitation is not critical, since the required buffer size $l$ remains small when the number of variables $n$ is low.

Once the buffer is filled, a new local model is trained using the stored samples. Specifically, the model parameters are obtained by minimizing the sum of squared errors over the buffered data $(\mathbf{X}_b, \mathbf{Y}_b)$. The matrix $\mathbf{X}_b \in \mathbb{R}^{l \times (n+1)}$ contains the $n$ input variables augmented with a column of ones to account for the bias term, while $\mathbf{Y}_b \in \mathbb{R}^l$ collects the corresponding output values. To cope with cases in which some of the variables do not change in the buffer data, we also add a L2 regularization parameter $\lambda$ (by default, $10^{-6}$). This parameter can also be optimized to avoid outliers when data is very noisy, as we will show in our experiments. Therefore, the parameters $\mathbf{w}$ of each local model is obtained using ridge regression as

\begin{equation}
 \mathbf{w} = (\mathbf{X_b}^T\mathbf{X_b} + \lambda \mathbf{I})^{-1}\mathbf{X_b}^T\mathbf{Y_b}.   
\end{equation}

In SyMPLER, the $i$-th local model is associated with an approximation point $\mathbf{x}=\mathbf{p}_i$, which defines the region of the input space for which the model is responsible. To justify how this point is chosen for each model, consider the hypothesis that the system has a true operating point $\mathbf{P}_i$ and that, during the collection of the buffer data, the input only fluctuates randomly around this point with zero-mean noise. Under this assumption, the sample mean $\hat{\mathbf{P}}_i$ of the buffer input data provides an unbiased estimate of the true operating point.

If these fluctuations are sufficiently small, Taylor’s theorem implies that the system behavior in the neighborhood of $\mathbf{P}_i$ can be accurately approximated by a linear model. In this context, the ordinary least-squares solution yields a consistent estimate of the coefficients of this local linear approximation. Since the regularized least-squares solution converges to the ordinary least-squares solution as $\lambda$ approaches 0, the parameters obtained by training a local model on the buffer data can be expected to approximate the coefficients of the Taylor expansion around $\hat{\mathbf{P}}_i$. Consequently, the approximation point of SyMPLER's $i$-th local model is defined as the mean of the buffer input data on which it was trained, that is, $\mathbf{p}_i = \hat{\mathbf{P}}_i$.

\end{revblock}



To ensure that the buffer data truly corresponds to a narrow region of the input space, the sampling period of the problem should be much smaller than the time it takes for the input to change its behavior. This is the case, for example, of many weather-related problems, in which the input data usually exhibits yearly periodic behavior and is sampled every hour or every minute. In the limit, when the sampling frequency is infinitely high, the buffer data converge to a single point in input space and the local model trained with it converges to the Taylor approximation around that point. In our experiments, we show that the coefficients of the hyperplanes obtained by SyMPLER around approximation points taken as the mean of the buffer data are very close to the true Taylor first-order coefficients around those points.

\subsection{VC-Theoretical Buffer Size}
\label{subsec:buffer}

The size of the buffer is a critical point for the model. In the one hand, if it is too small, the local model trained with it will likely overfit, specially if the data is noisy or has high dimensionality. In the other hand, if it is too big, it will cover a large range in the input space, for which the linearity assumption for the output will not hold. Therefore, we need the smallest buffer size that still ensures generalization of the local models beyond the training data.

Vapnik-Chervonenkis (VC) theory \cite{vapnik} provides generalization bounds for every model that chooses a set of parameters $\alpha$ based on data. For regression problems, these bounds usually take the form

\begin{equation}
    R(\alpha) \leq \frac{R_{emp}(\alpha)}{(1-c\sqrt{\varepsilon})_+},
\label{eq:risk_bound}
\end{equation}
where 
\begin{equation}
    \varepsilon = a_1\frac{h(\ln{\frac{a_2l}{h}}+1)-\ln(\frac{\eta}{4})}{l}.
    \label{eq:epsilon}
\end{equation}

\begin{revblock}

In Equations \eqref{eq:risk_bound} and \eqref{eq:epsilon}, $R(\alpha)$ is the risk of the model, defined as the expected value of its test error for a given parameter vector $\alpha$, whereas $R_{emp}(\alpha)$ denotes the empirical risk, corresponding to the training error for this parameterization. The term $h$ represents the VC dimension of the hypothesis class, $l$ is the size of the training set, and $c$ is a constant related to the tail probability of large errors. This inequality holds simultaneously for all $\alpha$ with probability $1-\eta$.

\end{revblock}

Cherkassky and Mulier \cite{cherkassky} suggest that, for most practical problems, it can be assumed that $a_1=a_2=c=1$. They use this bound to obtain the maximum capacity of regression models, measured with the VC dimension, for a fixed number of training samples. This follows from Eq. \eqref{eq:risk_bound}, since, for $\varepsilon \geq 1$, the expected risk of the model approaches infinity and the training error tells nothing about the test error. In this sense, thinking $\varepsilon$ as a function of $h$ with a fixed value of $l$, the maximum VC dimension $h^*$ a model can have is the solution of the following non-linear equation:

\begin{equation}
    \varepsilon(h^*) = \frac{h^*(\ln{\frac{l}{h^*}}+1)-\ln(\frac{\eta}{4})}{l} = 1
\label{eq:non-linear-eq}
\end{equation}

They solve Eq. \eqref{eq:non-linear-eq} using the bisection method and find that, for a large number of training samples, the solution is approximately $h^* = 0.8l$ for $\eta \geq \min(1, 4/\sqrt{l})$. For practical applications, however, they suggest $h \leq 0.5l$, since the bound \eqref{eq:risk_bound} can be too loose for some problems. 


The bounds provided by Cherkassky and Mulier can be used to set the smallest buffer size that ensures generalization in SyMPLER, since all the individual models in the network are linear and the exact VC-dimension for those models is known \cite{vapnik}. In the general case, a linear model of $n$ variables (not including the bias) has VC-dimension $h = n+1$. Hence, the smallest training set that guarantees generalization for this model, following from Cherkassky and Mulier, would be $l = 2(n+1)$.

However, since this approximation was obtained considering a large number of training samples, it is valid for models with large VC-dimension. For many practical regression problems, especially if some feature selection technique is applied, the linear models will have small VC-dimension and it will not hold anymore. For instance, for an univariate regression, the bound would suggest that a model could be trained without overfitting with only 4 samples, which is not practical. Therefore, the bound needs to be adjusted for small $h$. 

\begin{figure}[h]
    \centering
    \includegraphics[width=0.85\linewidth]{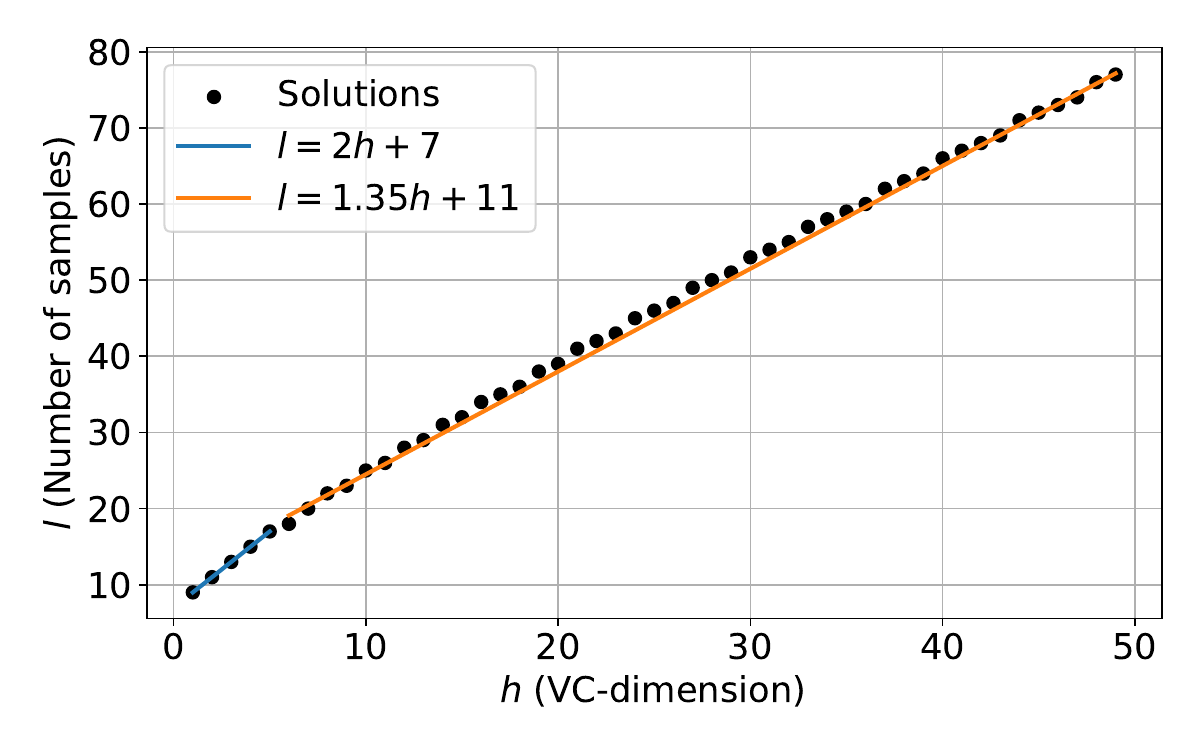}
    \caption{Smallest training size $l$ for a model with VC-dimension $h$, following \cite{cherkassky}.}
    \label{fig:buffer-bound}
\end{figure}

Following the same optimization procedure as Cherkassky and Mulier, but fixing the confidence level as $\eta = 0.01$, we find the solutions presented in Fig. \ref{fig:buffer-bound}. As expected, for large $h$ our solution is very similar to the original one. The line of best fit to the solutions with $h > 5$ is approximately $l = 1.35h + 11$, or $h = 0.74l - 8.14$, which is close to $h = 0.8l$ found originally. For smaller values of $h$, we find that $l = 2h + 7$, or $h = 0.5l - 3.5$, which is also similar to the ``practical" bound found by Cherkassky, $h \leq 0.5l$. However, the authors didn't include the offset term in their solution, which can't be neglected for smaller models. In fact, Fig. \ref{fig:buffer-bound} shows that if any model is trained with less than approximately 10 samples, there is no guarantee that it will not overfit. 

\begin{revblock}

Since these bounds mark the transition between finite and infinite generalization error, the buffer size must be strictly larger than them to guarantee generalization of the local models. A key component of the affine approximation of the solutions in Figure \ref{fig:buffer-bound} is the additive constant, which imposes a minimum training size below which no guarantees can be established. Its value depends on how the linear fit to the bound is constructed: in the low VC-dimension regime it is close to 7, whereas fitting over a wider range increases it to approximately 10.7. At the same time, the slope decreases from about 2 to roughly 1.4 as higher VC-dimensions are considered, reducing the minimum required number of samples. To obtain a robust guarantee across different model complexities, we adopt a conservative bound that combines the slope observed in the low VC-dimension regime with the larger additive constant arising in higher dimensions. Moreover, to simplify its practical use, we approximate the slope and the additive term by the rounded values of 2 and 10, respectively. Therefore, we propose that the buffer size for SyMPLER in a problem with $n$ variables should be 

\begin{equation}
    l = 2(n+1) + 10,
\label{eq:final-bound}
\end{equation}
which allows us to train local models without overfitting.

Our preference for this conservative bound is further motivated by the temporal dependence inherent in time-series data. The VC-based bounds rely on the assumption that samples are independent and identically distributed (i.i.d.), which must be treated with caution in longitudinal settings. Assuming that the sampling period is much shorter than the time scale of regime changes effectively corresponds to stationarity within each training buffer, satisfying the identically distributed requirement. As discussed previously, this assumption is reasonable in several practical applications, such as weather-related problems, where measurements are collected frequently while regime changes occur over much longer time scales.

Even under stationarity, however, the independence assumption is typically violated due to temporal correlations. As a result, the effective number of independent samples used to train each model is smaller than the number of observations stored in the buffer, with stronger dependence leading to larger discrepancies. In practice, this issue is mitigated by the fact that VC-based generalization bounds are known to be highly conservative, as they are derived under worst-case assumptions and often prescribe training sizes larger than the minimum required in practice \cite{uniform-convergence-dl}. Adopting a conservative affine approximation to the solutions in Fig. \ref{fig:buffer-bound} therefore increases the safety margin with respect to the effective sample size and strengthens generalization guarantees in the presence of temporal dependence.


\end{revblock}

\subsection{Switching between Local Models}
\label{subsec:switch}

\begin{revblock}
    
From Eq. \eqref{eq:taylor-multivariate}, the approximation error of a first-order Taylor expansion at a test point $\mathbf{x}$ depends on the Euclidean distance $D = \lVert \mathbf{x} - \mathbf{p}_0 \rVert_2$ between $\mathbf{x}$ and the approximation point $\mathbf{p}_0$. This reflects the intuitive fact that a local model provides accurate predictions in the vicinity of its approximation point and becomes less accurate as one moves away from it.


Based on this observation, model selection in SyMPLER is performed by choosing, for each test input $\mathbf{x}$, the local model whose approximation point is closest to $\mathbf{x}$. Specifically, given the set of approximation points $\{\mathbf{p}_i\}_{i=1}^m$, the selected model index is defined as

\begin{equation}
    i^*(\mathbf{x}) = \arg \min_{i \in \{1,\dots, m\}}||\mathbf{x}-\mathbf{p}_i||
    \label{eq:switching-criterion}
\end{equation}

Although the error also depends on a higher order derivative of the real function at $\mathbf{p}_0$, this information is generally unavailable. Nevertheless, we will show that the proposed distance-based criterion is sufficient for choosing between the models and provides SyMPLER with strong approximation capabilities.

\end{revblock}

\section{Experiments and Discussion}
\label{sec:exps}



This section presents the application of SyMPLER to two nonlinear dynamical systems. First, the method is trained online in a simulated environment to model the acceleration of an undamped pendulum based on its angular position and then used for long-term forecasting. Next, SyMPLER predicts hourly electric load of a real US utility from temperature data, with results compared to other continual learning methods using fitting error, prediction error, and forgetting rate metrics.

\subsection{Simulation Data}
\label{subsec:simulation}

To illustrate the proposed method, it was applied to a classical system identification task: modeling the nonlinear dynamics of the simple undamped pendulum, governed by the equation

\begin{equation}
    \Ddot{\theta} = -\frac{g}{l}\sin\theta,
\label{eq:pendulum}
\end{equation}
where $\theta$ is the angular position of the pendulum in relation to the vertical axis, $\Ddot\theta$ is its angular acceleration, $l$ is the length of its rod and $g=9.81$ m/s² is the acceleration of gravity.



Usually, to perform a linear analysis on the behavior of this system, it is assumed that its angular position is always close to zero, so that $\sin\theta \approx \theta$ and Equation (\ref{eq:pendulum}) becomes linear. However, in practical scenarios this assumption -- and other stationarity assumptions used to linearize systems around equilibrium points -- do not hold and the predictions of the model become worse as it drifts away from the assumed operation point. To overcome this limitation, while still maintaining locally-linear interpretability of the system, SyMPLER can be used to model its behavior.

In this context, the pendulum was simulated with a rod length of $l=0.5$ m starting at the horizontal position ($\theta = 90$°) and its position was sampled with a rate of 200 Hz throughout four cycles of its operation. At each instant $t$, its position was obtained using Runge-Kutta 4th order method and its current velocity and past acceleration were estimated using finite difference method,
simulating the practical scenario where we don't have direct access to these variables. The objective of the model is to estimate the current acceleration of the pendulum given its current position. However, the acceleration depends on future information about the position of the system, so it becomes a time series forecasting problem in which we use $\theta(t)$ to predict the estimate of $\Ddot{\theta}(t)$, which will only be available after $dt$ seconds. Since the acceleration depends non-linearly on the input, the model needs to adapt itself continuously to model the input-output relation though all of the input's range, characterizing a continual learning problem.

\begin{figure}[h!]
    \centering
    \includegraphics[width=\linewidth]{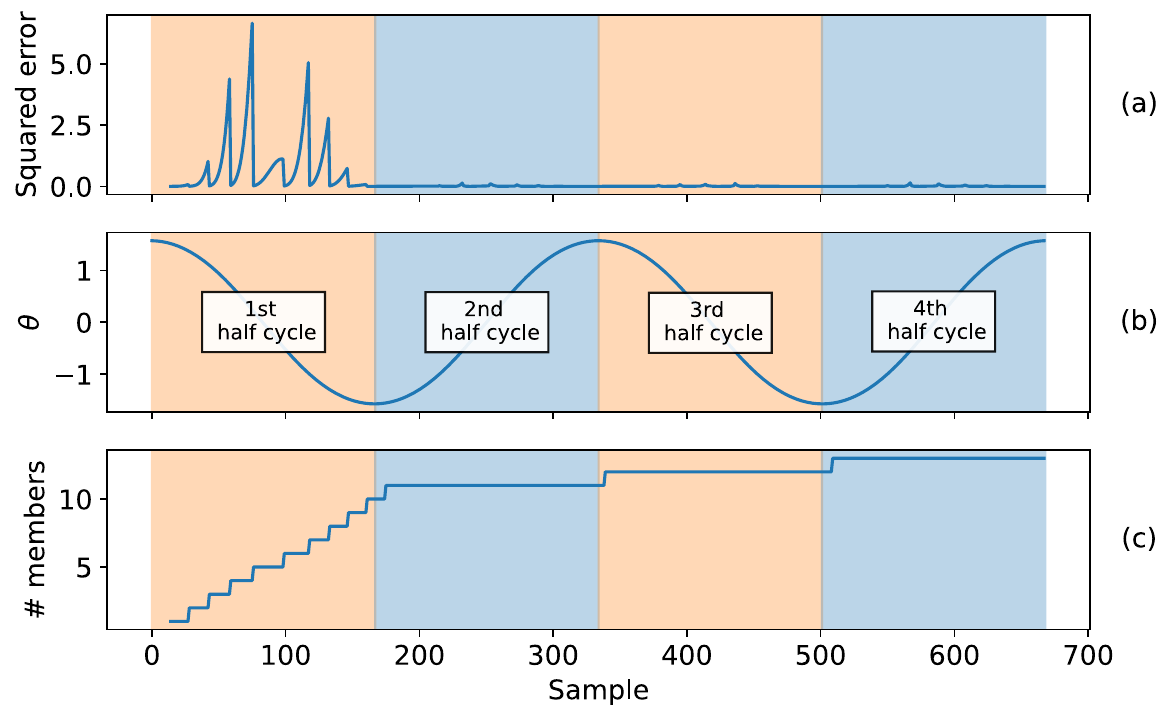}
    \caption{Continual learning results: (a) prediction squared error, (b) input data and (c) number of members in the network. Data in (a) and (c) start appearing later than in (b) because the network has no member until the first $2(n+1)+10$ samples arrive.}
    \label{fig:pendulum-train}
\end{figure}

The first two cycles of the collected data were used to train the model. As shown in Figure \ref{fig:pendulum-train}, many members were added to the network during the first half cycle of the input. As the network had not modeled any region of the data at that point, the prediction error of the model became larger than naïve's and the evolution criteria was satisfied. During the other three half cycles, the model error was much lower -- since it had already learned the behavior of the output for positive and negative values of the input -- and fewer models were added. SyMPLER's error remained low during all of the remaining training, since it benefited from learning the desired function for all of the input's range. Still, some members were added to the network when the input values were close to the peaks and valleys of the input. This is due to the naïve model's performance being favoured in this region, since the value of the acceleration remains approximately constant on them.

\begin{figure}[h!]
    \centering
    \includegraphics[width=0.82\linewidth]{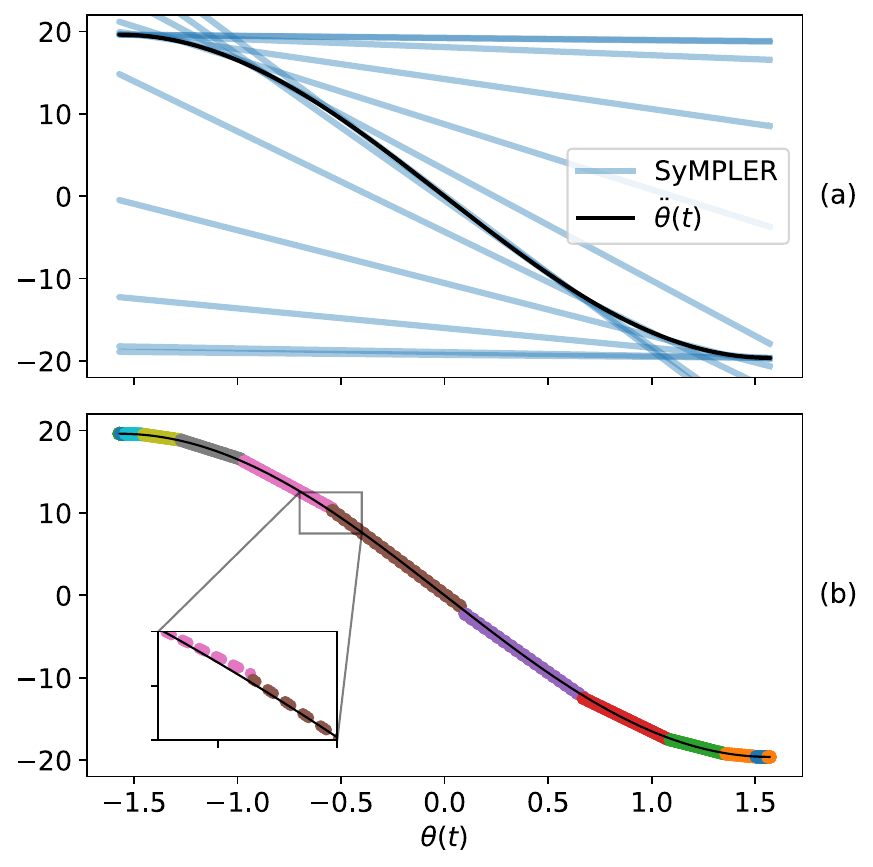}
    \caption{SyMPLER approximation in feature space. Although each individual model is linear, as shown with the coloured lines in (a), switching between them allows the network to approximate non-linear functions piecewise-linearly, as shown in (b). The zoomed area shows the piecewise-linear behavior of the model.}
    \label{fig:pendulum-lines}
\end{figure}

In total, 13 local models were added to the network. In the feature space, each one of these models is a line that approximates tangents to the graph of $\Ddot{\theta} = -\frac{g}{l}\sin\theta$. Figure \ref{fig:pendulum-lines} (a) shows the response of the individual models for test data composed of two cycles of the input. Although each model only approximates the target function in a small region, switching between the models allows the network to approximate it in its full range. In this sense, Figure \ref{fig:pendulum-lines} (b) shows the final response of the network. The zoomed region shows the piecewise linear behavior of the model. 


To check the assumption that SyMPLER's individual models represent first-order Taylor approximations of the target function, the parameters of the individual models are compared to the Taylor approximations obtained analytically. Since the generator function is known from Equation (\ref{eq:pendulum}), the first-order Taylor polynomial that approximates it near a point $\theta_0$ can be obtained using Equation (\ref{eq:taylor-univariate}). Table \ref{tab:taylor-parms} shows the coefficients of the analytical Taylor approximations and the coefficients of the committee members for some of the approximation points. As supposed, the individual models approximate the Taylor first-order expansions. 



\begin{table}[h]
    \centering
    \renewcommand{\arraystretch}{1.2} 
    \setlength{\tabcolsep}{10pt} 
    \caption{Coefficients of Taylor first-order approximations of the target function and coefficients of SyMPLER members for some of the approximation points.}
    \label{tab:comparison}
    \begin{tabular}{cccc}
        \toprule
        $\theta_0$ (rad)&       & Slope & Bias \\
        \midrule
        \multirow{2}{*}{1.55} & SyMPLER & -0.44 & -18.91 \\
                                & Taylor  & -0.34 & -19.07 \\
        \midrule
        \multirow{2}{*}{1.24} & SyMPLER & -6.42 & -10.54 \\
                                & Taylor  & -6.34 & -10.69 \\
        \midrule
        \multirow{2}{*}{0.44} & SyMPLER & -17.72 & -0.49 \\
                                & Taylor  & -17.78 & -0.53 \\
        \midrule
        \multirow{2}{*}{-1.52} & SyMPLER & -0.97 & 18.10 \\
                                & Taylor  & -0.88 & 18.25 \\
        \bottomrule
    \end{tabular}
\label{tab:taylor-parms}
\end{table}

\begin{figure}[h!]
    \centering
    \includegraphics[width=\linewidth]{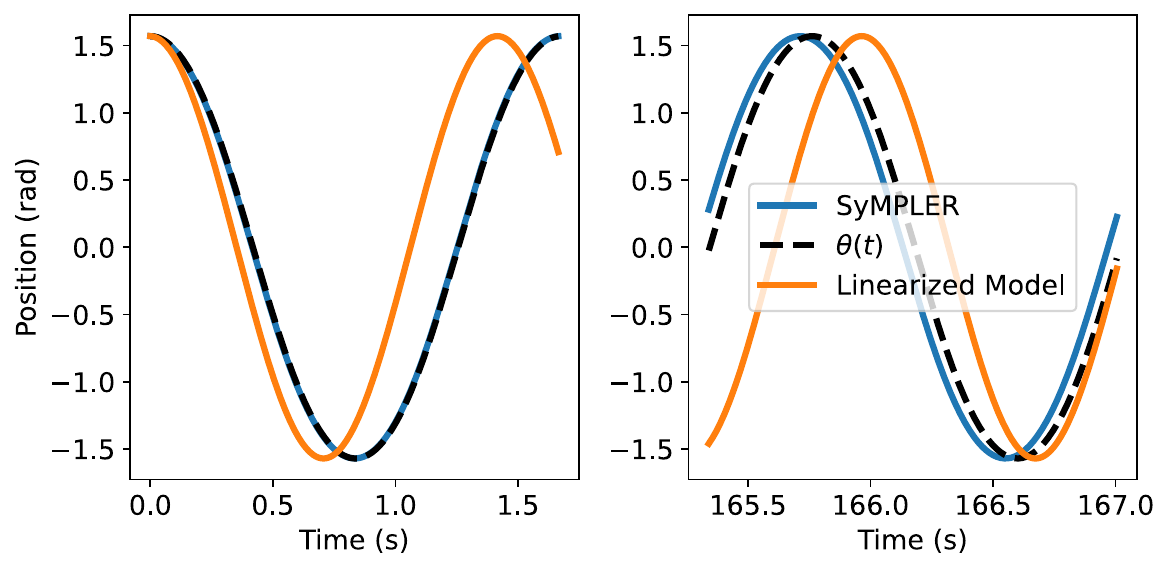}
    \caption{Long-term forecasting of the angular position of the pendulum using SyMPLER and the linearized model for the first cycle of the system (left figure) and its last cycle (right figure).}
    \label{fig:pendulum-forecasting}
\end{figure}

This model was fixed and used to perform a long-term forecasting of the pendulum angle based only on the information about its starting position ($\theta = \frac{\pi}{2}$ rad = 90°) and velocity ($\dot{\theta} = 0$ rad/s). At each step, the model uses the position information (predicted on the previous step) to estimate the current acceleration, which is then used to predict the next position. This procedure was held for 167 seconds, corresponding to approximately 100 cycles of the pendulum. Figure \ref{fig:pendulum-forecasting} shows the predicted angular position of the pendulum during its first two cycles and last two cycles for SyMPLER and the model linearized around $\theta=0$°. Naturally, as SyMPLER provides more accurate predictions for the acceleration of the pendulum for the full range of its positions, its forecasts follow the true position of the pendulum accurately for much longer than the linearized model's. In fact, the latter rapidly loses track of the angular position of the pendulum, while SyMPLER's errors accumulate much slowly, providing good results even for the last cycles.

\subsection{Real Data}
\label{subsec:real}


Our methodology was applied to an electric load forecasting problem in order to evaluate its performance in a realistic setting and to compare it against state-of-the-art approaches. We use the dataset from the Global Energy Forecasting Competition of 2014 (GEFCom2014 \cite{gefcom}), which focuses on the construction of probabilistic models to predict the electrical demand of a U.S. utility based on hourly temperature measurements collected from 25 weather stations. SyMPLER is compared against a neural network trained using EWC-SWFT \cite{ewc-based}, an evolving Takagi–Sugeno (eTS) fuzzy model \cite{ets}, \rev{and Incremental Controller Networks (ICN) \cite{icn}}, as well as two offline baselines: a nonlinear model (XGBoost) and a linear regression model. \rev{To enable a fair comparison in an online setting, a slight adaptation of the original ICN was required. Specifically, when the criterion for adding a new model is met, instead of halting the process and sampling the system at the current operating point (as assumed in the original formulation), the new model is initialized with a single sample and updated as new data arrive, following ICN's update rule.}

\rev{In addition, we also compare the original SyMPLER formulation with a slightly modified version to allow model interpolation. While the method originally employs a winner-take-all strategy for model selection, this may lead to discontinuities in the predictions. Therefore, we include in our experiment an alternative aggregation scheme in which multiple local models contribute to the output. In this variant, model contributions are weighted by an exponential decay function of their distance to the current operating point, followed by normalization to form a convex combination. This aggregation, which resembles fuzzy local models' interpolation, introduces an additional hyperparameter, the decay rate $\sigma$ of the exponential function, which is optimized jointly with the regularization coefficient.}

The data spans from 2005 to 2011. However, since it exhibits yearly periodicity, we use only data from 2005, 2006 and 2007 for, respectively, feature and hyperparameters selection, continual training and forecasting. The training and forecasting data were split following CLeaR's evaluation setup \cite{clear}, which suggested measuring three metrics for assessing the performance of a continual learning method: fitting error, prediction error and forgetting ratio. To do so, the data has to be split into three sequential subsets: a first small subset to fit the model to one region of the data (called warmup phase), a second larger subset to model the full range of the input (called update phase) and a third subset to make predictions with the model fixed (called evaluation phase). The three metrics are evaluated as follows:

\begin{itemize}
    \item Fitting error: root mean squared error (RMSE) of the model on forecasting warmup and update data after it was trained on both. It measures the model's ability to fit the training data.
    \item Prediction error: forecasting RMSE of the model on data from the evaluating phase. It measures the model's performance outside of training data.
    \item Forgetting ratio: measures model's ability to remember warmup data after it was trained on update data. A continual learning model should be able to remember past behavior of the problem even after being updated on new behavior. This metric is calculated as
    \begin{equation}
        \text{forgetting ratio} = \frac{\max (0, L_{\text{w}}^{\text{u}}-L_{\text{w}}^{\text{w}})}{L_{\text{w}}^{\text{w}}},
    \end{equation}
    where $L_{\text{w}}^{\text{w}}$ is the forecasting error on the warmup data after the warmup phase and $L_{\text{w}}^{\text{u}}$ is the error on the warmup data after the update phase.
\end{itemize}

Following this procedure, we selected the data from January of 2006 for the warmup phase (744 samples), the remaining data from this year for the update phase (8016 samples) and all the data from 2007 for the evaluation phase (8760 samples).


Before training, we apply a multivariate mutual information-based feature selection \cite{kraskov} on 2005 data. This is a very important step for continual learning based on local models, such as SyMPLER and Takagi-Sugeno approaches, since they can suffer from the curse of dimensionality \cite{survey-fuzzy}. \rev{(The effect of the input dimensionality on SyMPLER's behavior is addressed experimentally in Section \ref{appendix-highdim})}. To make the feature selection computationally tractable, as well as robust to the data variance, we apply the method on five bootstrap subsets of 1000 samples and select the variables that are selected at least twice. From the 25 initial variables, each one corresponding to the temperature measured on a different weather station one hour before the forecast, we extract the 8 more informative and less redundant among them. Namely, weather stations 1, 6, 11, 14, 20, 21, 22, and 25 were were selected.

The same data was used to optimize the hyperparameters of the models. Half of the data was used to train the models and the other half was use to test them, without updating. A grid search was performed to select the hyperparameters combination that provided the lowest MSE for each model on the test data. In this context, Table \ref{tab:hyperparms} shows the results of the optimization procedure. Since the lower bound obtained for the number of training examples for each model of SyMPLER represents a threshold between finite and infinite test risk, it can be useful to regularize the models, specially when dealing with very noisy data, as it is the case. \rev{In Section \ref{appendix-noisy}, we explore experimentally the role regularization plays on the quality of SyMPLER's local approximations in the presence of different levels of noise in the data.}
%

\begin{table}[h]
    \centering
    \renewcommand{\arraystretch}{1.2}
    \setlength{\tabcolsep}{3pt}
    \caption{Hyperparameter search and best parameters for each model.}
    \label{tab:hyperparms}
    \begin{tabular}{@{}lll@{}}
        \toprule
        \multicolumn{1}{c}{Model} &
        \multicolumn{1}{c}{Searched parameters} &
        \multicolumn{1}{c}{Best parameters} \\
        \midrule
        \begin{tabular}[c]{@{}l@{}}
            SyMPLER\\
            (original)
        \end{tabular} &
        $\lambda = \{10^{-6}, 10^{-2}, 0.1, 1, 5, 10, 15\}$ &
        $\lambda = 5$ \\
        \midrule
        \begin{tabular}[c]{@{}l@{}}
            \rev{SyMPLER}\\
            \rev{(aggr.)}
        \end{tabular} &
        \begin{tabular}[c]{@{}l@{}}
            \rev{$\lambda = \{10^{-6}, 10^{-2}, 0.1, 1, 5, 10, 15\}$} \\
            \rev{$\sigma = \{0.1, 0.5, 1, 10\}$}
        \end{tabular} &
        \begin{tabular}[c]{@{}l@{}}
            \rev{$\lambda = 1$} \\
            \rev{$\sigma = 1$}
        \end{tabular} \\
        \midrule
        \rev{ICN} &
        \begin{tabular}[c]{@{}l@{}}
            \rev{$\lambda = \{10^{-6}, 0.1, 1, 5\}$} \\
            \rev{$\epsilon = \{10^{-2}, 0.5, 1, 5\}$} \\
            \rev{$\alpha = \{0.1, 1, 10\}$}
        \end{tabular} &
        \begin{tabular}[c]{@{}l@{}}
            \rev{$\lambda = 5$} \\
            \rev{$\epsilon = 10^{-2}$} \\
            \rev{$\alpha = 0.1$}
        \end{tabular} \\
        \midrule
        eTS &
        \begin{tabular}[c]{@{}l@{}}
            $\omega = \{10^{-6}, 0.1, 1, 10, 100, 1000\}$ \\
            $r = \{0.1, 0.3, 0.5, 0.7, 0.9\}$
        \end{tabular} &
        \begin{tabular}[c]{@{}l@{}}
            $\omega = 0.1$ \\
            $r = 0.5$
        \end{tabular} \\
        \midrule
        EWC-SWFT &
        \begin{tabular}[c]{@{}l@{}}
            $\beta = \{100, 500, 1000\}$ \\
            $\text{learning\_rate} = \{10^{-4}, 10^{-3}, 10^{-2}\}$
        \end{tabular} &
        \begin{tabular}[c]{@{}l@{}}
            $\beta = 100$ \\
            $\text{learning\_rate} = 10^{-3}$
        \end{tabular} \\
        \midrule
        XGBoost &
        \begin{tabular}[c]{@{}l@{}}
            $\text{n\_estimators} = \{10, 100, 200, 500\}$ \\
            $\text{max\_depth} = \{3, 6, 9\}$ \\
            $\text{learning\_rate} = \{0.01, 0.1, 0.3\}$ \\
            $\text{colsample\_bytree} = \{0.3, 0.6, 1\}$ \\
            $\text{alpha} = \{0, 10, 100\}$
        \end{tabular} &
        \begin{tabular}[c]{@{}l@{}}
            $\text{n\_estimators} = 200$ \\
            $\text{max\_depth} = 3$ \\
            $\text{learning\_rate} = 0.1$ \\
            $\text{colsample\_bytree} = 0.3$ \\
            $\text{alpha} = 0$
        \end{tabular} \\
        \bottomrule
    \end{tabular}
\end{table}

%
All data -- warmup, update, and evaluation -- were standardized using the mean and standard deviation from 2005. As with other EWC-based continual learning algorithms, EWC-SWFT requires defining a task that the model should retain while continuously adapting. In our setup, the warmup data served as this task, since the model’s forgetting factor is computed from it. During the update phase, a one-month sliding window was used, initialized with the warmup data. The model was then trained for 400 epochs on the full window, advancing one sample at a time. SyMPLER, eTS and ICN were trained continuously both on warmup and evaluation phases, with samples arriving in a stream. In contrast, the offline baselines were trained using the entire warmup dataset, followed by the full update dataset. As these models do not implement any mechanism to prevent forgetting and have access to the complete dataset at once, both should serve as worst-case baselines for forgetting ratio and XGBoost should also serve as a best-case baseline for fitting and prediction errors. 

\begin{revblock}
    
Table \ref{tab:results} summarizes the experimental results. Using 291 and 286 local models, respectively, both the original version of SyMPLER and the version with model aggregation achieved superior performance across all evaluation criteria when compared to eTS, which instantiated only 14 local models. This contrast suggests that eTS added fewer local models than required to adequately capture the nonlinear and nonstationary structure of the problem. In comparison, ICN added 8,680 local models, nearly matching the total number of training samples (8,760), indicating an excessive model proliferation. As a consequence, ICN was unable to update its local models with a sufficient number of samples to ensure generalization. It should be noted that this behavior partially reflects the adaptation of ICN to online forecasting, as the original formulation assumes that the process can be halted for active data collection, an assumption not applicable in this setting. Despite this limitation, ICN achieved a competitive forgetting ratio, outperforming eTS and EWC-SWFT. Both SyMPLER variants achieved prediction errors comparable to XGBoost, a state-of-the-art offline regression method, while exhibiting substantially lower forgetting, since XGBoost lacks mechanisms to prevent catastrophic forgetting. The poor performance of the linear model further confirms the nonlinear and nonstationary nature of the problem. EWC-SWFT presented the highest prediction and fitting errors, despite a low forgetting factor, due to its task being defined as the warmup data. As a result, forgetting of previously acquired knowledge during the update phase is not penalized, causing the model to favor recent data outside the task window. In fact, this behavior leads to low errors near January and December (corresponding to the period of the task and the final sliding window, respectively) but large degradation over the rest of the year, resulting in the highest overall error. More generally, EWC-SWFT requires explicit task specification and repeated recalculation of penalization terms using the full historical dataset, which limits its practicality in streaming scenarios. Overall, SyMPLER consistently outperformed the other online methods: while the winner-take-all version achieved the lowest forgetting ratio, the aggregated variant provided superior predictive accuracy with comparable forgetting, yielding the best overall trade-off for this problem.

\end{revblock}

\begin{figure*}[t] 
    \centering
        \includegraphics[width=0.95\textwidth]{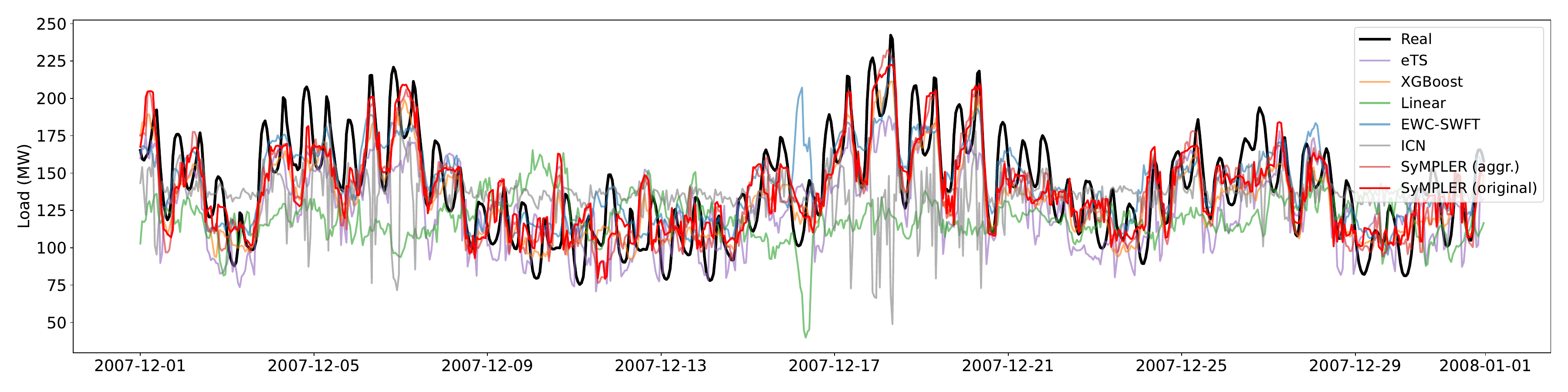} 
    \caption{Predictions of the models for the last month of 2007.}
    \label{fig:predictions}
\end{figure*}

Figure \ref{fig:predictions} shows the models' predictions for the final month of the evaluation subset. The linear offline model clearly performs the worst in this case, predicting values close to the mean of the response variable. The other models perform much better, predicting the electrical load accurately, especially if we consider that the models do not use auto-regressive variables for predicting the output. As argued, EWC-SWFT yields good predictions for December data, since it corresponds to the period covered by the final window it was trained on. In this region, eTS seems to often underestimate the load. Considering that it got a higher fitting error than SyMPLER, this indicates that the 14 models it placed were not enough to fully approximate the nonlinear dependencies of the problem. In this sense, the larger number of local models created by SyMPLER may be justified, since it lowered both the fitting and the prediction error without causing catastrophic forgetting.


\begin{table}[h]
    \centering
    \caption{Comparison of models based on fitting error (RMSE), prediction error (RMSE), and forgetting ratio. EWC-SWFT performance was evaluated for 10 random initializations, so we present the mean and standard deviation of the results.}
    \begin{tabular}{lccc}
        \toprule
        Model & \begin{tabular}[c]{@{}c@{}}
            Fitting\\
            error
        \end{tabular} & \begin{tabular}[c]{@{}c@{}}
            Prediction\\
            error
        \end{tabular} & \begin{tabular}[c]{@{}c@{}}
            Forgetting\\
            ratio
        \end{tabular} \\
        \midrule
        SyMPLER (original) & 21.91  & 21.06  & \textbf{0.11}  \\
        \rev{SyMPLER (aggr.)} & \rev{\textbf{21.09}} & \rev{\textbf{21.01}} & \rev{0.14} \\
        \rev{ICN} & \rev{31.03} & \rev{39.07} & \rev{0.15} \\
        eTS      & 26.88  & 28.87  & 0.30  \\
        EWC-SWFT & 49.10 $\pm$ 3.86 & 55.64 $\pm$ 4.45 & 0.25 $\pm$ 0.29 \\
        \midrule
        XGBoost  & \textbf{16.88}  & \textbf{18.56}  & \textbf{0.66}  \\
        Linear   & 39.12  & 44.19  & 1.06  \\
        \bottomrule
    \end{tabular}
    \label{tab:results}
\end{table}

Although the evolving Takagi-Sugeno approach yielded a model easier to interpret, the low number of local models added negatively impacted its performance. Since the models don't use the past values of the output variable to predict it, SyMPLER added much more local models trying to follow the performance of the naïve predictor, which is usually very competitive and robust to nonstationarity for time series that vary slowly. On the one hand, this negatively impacts the model's interpretability, since it is explained by more rules; on the other hand, this substantially improved the model's performance, approaching a state-of-the-art model trained offline. Furthermore, SyMPLER's predictions remain fully explainable, since only the local model associated to the approximation point closest to a test example is used to predict its output. Analyzing the weights of this local model, one can infer the behavior of the electrical demand for small variations around the temperature values of that point. For instance, see Table \ref{tab:interpret}. It illustrates two scenarios: in the first, all the weather stations measure 0°C (or 32°F); in the second, all measure 30°C (or 86°F). For simplicity, the analyses were performed over the original SyMPLER model (without aggregation of local models), since only one model responds to each input pattern.

Consider the first scenario, represented by $x_{\text{0}}$. The closest approximation point to it in the trained model is $p_{\text{0}}$, to which is associated a local model with weights $w_{\text{0}}$ considering the original data (not standardized) in Fahrenheit and the last weight being the bias. This local model tells us that, if the temperature measured by the stations increases by a small amount, the electrical demand will decrease in the next hour, since most of the weights are negative. Conversely, if the temperature decreases, the expected electrical demand will increase. This result makes intuitive sense, since this is the scenario of a cold day and increasing the temperature would lower the electric demand of heaters, for instance, and decreasing the temperature would increase it. 

The results for the second scenario are also intuitive. Since $x_1$ represents a scenario of a hot day, increasing the temperature will, in average, increase the electrical demand because of cooling systems, for example. Accordingly, decreasing the temperature decreases the load, since the cooling systems would have less demand. These behaviors are confirmed by the weights of the local model associated to this region being mostly positive, indicating that this interpretation is valid for small temperature changes around $x_1$.

\begin{table}[h]
    \centering
    \caption{The local approximation points closer to the scenarios $x_{\text{0}}$ and $x_{\text{1}}$ in which all the weather stations measure 32°F and 86°F, respectively, are $p_{\text{0}}$ and $p_{\text{1}}$. The local models associated with these points have weights $w_{\text{0}}$ and $w_{\text{1}}$.}
    \begin{tabular}{lccc|ccc}
        \toprule
        Station & \( x_{\text{0}} \) (°F) & \( p_{\text{0}} \) (°F) & \( w_{\text{0}} \) & \( x_{\text{1}} \) (°F) & \( p_{\text{1}} \) (°F) & \( w_{\text{1}} \) \\
        \midrule
        14 & 32.00 & 37.11 & 1.48  & 86.00 & 83.93 & -12.30 \\
        6  & 32.00 & 30.18 & -4.75 & 86.00 & 85.64 & 8.57  \\
        1  & 32.00 & 30.82 & -8.53 & 86.00 & 83.82 & 19.63 \\
        25 & 32.00 & 28.54 & -7.54 & 86.00 & 83.18 & 10.20 \\
        21 & 32.00 & 31.86 & -2.27 & 86.00 & 89.29 & 13.10 \\
        20 & 32.00 & 29.11 & -8.69 & 86.00 & 83.93 & 12.64 \\
        11 & 32.00 & 31.86 & -6.45 & 86.00 & 85.36 & 15.24 \\
        22 & 32.00 & 31.57 & -3.31 & 86.00 & 86.54 & 4.20  \\
        Bias      & --    & --    & 133.70 & --    & --    & 117.69 \\
        \bottomrule
    \end{tabular}
    \label{tab:interpret}
\end{table}

Despite SyMPLER's strong results in both experiments, future improvements are possible, especially regarding its ability to cope with concept drifts -- not only covariate drifts. In this case, since the generator function will have different behaviors through time for a same region of the input, there will be several local models for each region. Then, using only the distance to the approximation points will not be conclusive as to which model should be selected. Other criteria, different from the Euclidean distance, can be used to select the local models at each instant. For instance, using the past error as a tiebreaker, the most up-to-date model can be selected without discarding the past knowledge about the problem. \rev{In Section \ref{appendix-concept}, this alternative selection criteria is experimentally validated in data with the presence of concept drifts.}

Furthermore, different problems may demand different naïve baselines to trigger SyMPLER updates. In our experiments, we used a delayed predictor, which assumes that the next state of the output is similar to the last. This approach is effective for many real-world forecasting tasks, particularly short-term predictions where the system evolves gradually. However, for problems with weaker auto-regressive dependencies, a more suitable baseline may be needed, although selecting one is not always straightforward. One possibility that may fit the general case is using other continual learning methods -- such as black-box models -- as reference predictors, allowing SyMPLER to approximate their performance while maintaining interpretability. However, this introduces additional computational costs, and determining the optimal baseline for general use remains an open question.

SyMPLER's adaptive structure, combining plasticity and the ability to remember the behavior of the problem in the past, could also be adapted to learn high-dimensional problems. The main current limitation to this is the loss of locality in higher dimensions, leading to the curse of dimensionality. As previously stated \rev{(and detailed in Section \ref{appendix-highdim})}, this makes it essential to perform feature selection in order to reduce its dimensionality and maintain the local behaviors of the problem. However, if low-dimensional features are extracted before training SyMPLER -- using a neural network, for instance --, it could be promptly trained in an online fashion, adapting to drifts in feature space. Although the model could lose its interpretability, depending if the extracted features are interpretable, its training procedure would be the same, since the VC-theoretical bounds also hold for the feature space.

\begin{revblock}

\section{Additional Experiments}
\label{sec:add-exp}

\subsection{Experiments with Concept Drifts}
\label{appendix-concept}

To evaluate SyMPLER under concept drift, which causes newly added local models to overlap with obsolete ones in the input space, we test an alternative model selection strategy based on recent prediction error, replacing the original distance-based criterion. Under this rule, the local model with the smallest prediction error on the most recent sample is selected to predict the next output. The experiment is conducted adapting the pendulum benchmark by inducing a change in the system dynamics through a variation of the rod length from $l = 0.5$ m to $l = 1$ m during training. SyMPLER is trained continuously over two oscillation periods spanning both dynamics, using the proposed error-based criterion for model selection.

Figure~\ref{fig:pendulum_concept_error_and_members} shows the prediction error, number of local models, and input signal during training. During the first training cycle, the model behaves as in the original experiment, since only covariate shifts are present. When the system dynamics change, a sharp increase in prediction error triggers the addition of new local models operating over the same input domain as earlier ones. After sufficient models have been introduced to capture the new dynamics, the prediction error decreases again and model addition ceases. This indicates that, despite the coexistence of obsolete and newly added models in the same region of the input space, the proposed error-based selection criterion successfully prioritizes the most up-to-date models, yielding accurate predictions. However, since the selected local model may not be the one trained closest to the current operating point, its parameters may not represent a strictly local approximation and should therefore be interpreted with care.

\begin{figure}
    \centering
        \includegraphics[width=\linewidth]{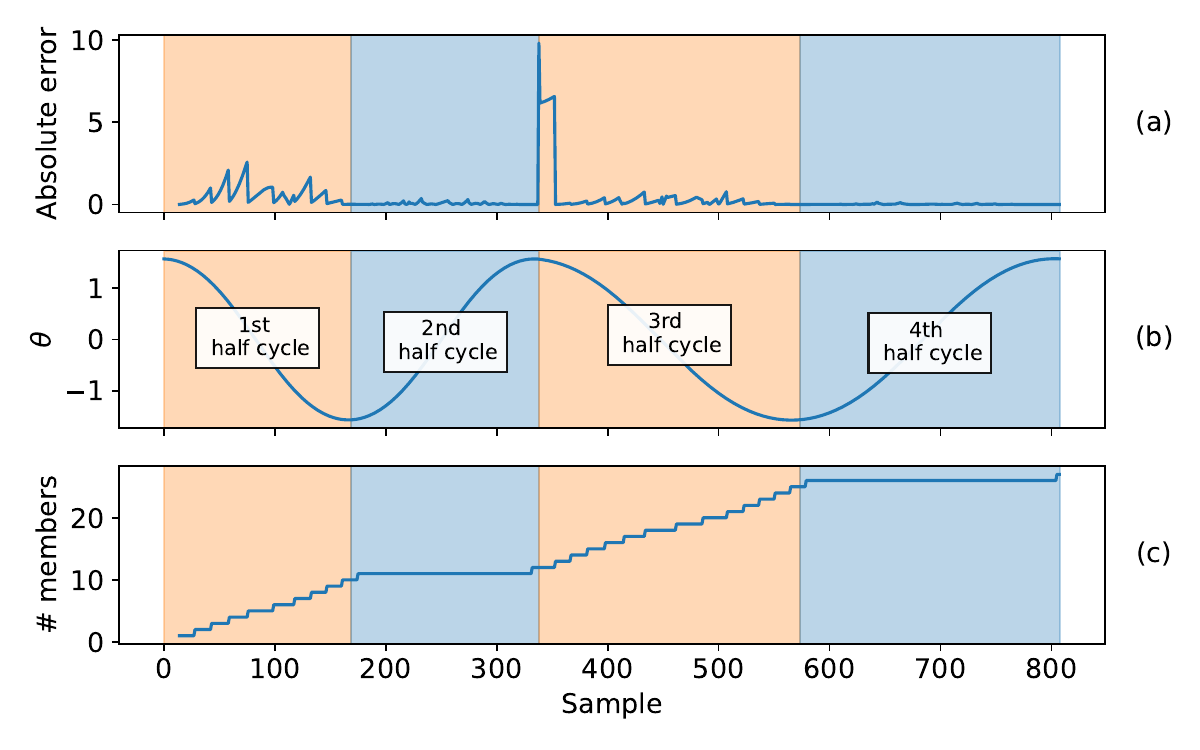}
    \caption{\rev{Continual learning results for the concept drift experiment: (a) absolute prediction error, (b) input data and (c) number of members in the network.}}
    \label{fig:pendulum_concept_error_and_members}
\end{figure}

\subsection{Experiments with High-Dimensional Data}
\label{appendix-highdim}

To assess the behavior of SyMPLER as input dimensionality increases, we extend the pendulum experiment by progressively augmenting the data with spurious variables sampled from a standard normal distribution. The number of added variables ranges from 0 to 100. For each number of spurious variables, SyMPLER is trained continuously over two oscillation periods and evaluated on a subsequent period. This process is repeated 10 times to account for variability.

Figure \ref{fig:pendulum_high_dim_mse_and_members} shows the test MSE and the number of local models as functions of the input dimensionality. As noisy variables are introduced, the number of local models initially increases due to degraded prediction accuracy, which triggers model creation. As dimensionality increases further, the minimum number of samples required to train a local model grows with the input dimension (Eq. \ref{eq:final-bound}), suppressing model creation even when prediction errors remain high. In this regime, only a few effectively global models cover the input space, leading to increased error due to both noise and loss of locality.

\begin{figure}
    \centering
        \includegraphics[width=0.84\linewidth]{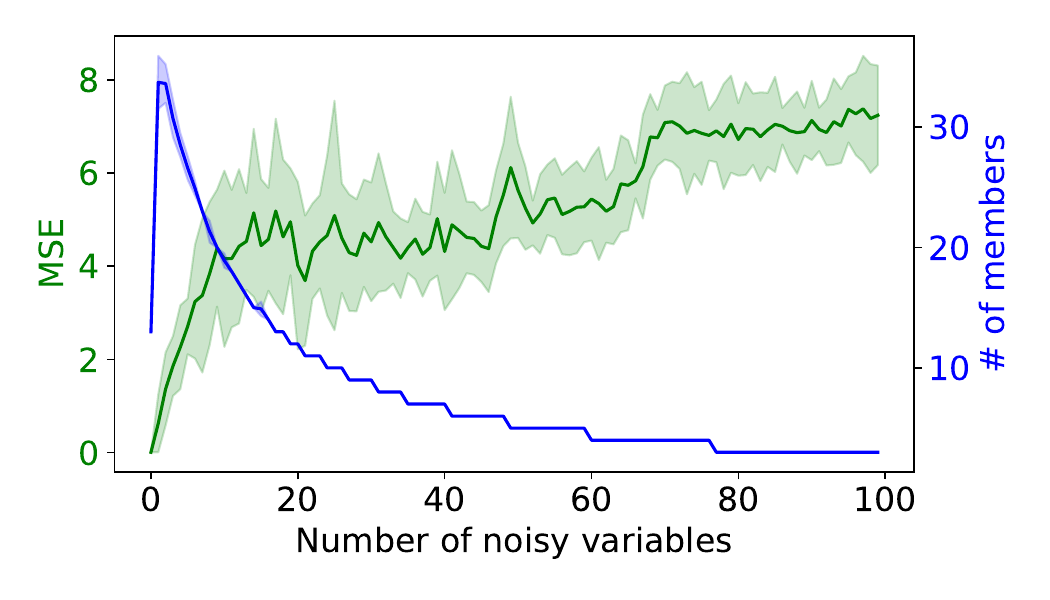}
    \caption{\rev{SyMPLER's average error and number of members for an increasing input dimensionality. The shaded areas represent two standard deviations from the mean.}}
    \label{fig:pendulum_high_dim_mse_and_members}
\end{figure}

%
%
Although these results represent a worst-case scenario in which the additional variables carry no information about the output, they emphasize the importance of feature selection when dealing with local models.

\subsection{Experiments with Noisy Data}
\label{appendix-noisy}

To assess the effect of noise on SyMPLER’s local approximations, we extend the pendulum experiment by adding increasing levels of noise to the output and comparing the learned local model parameters to the analytical Taylor coefficients of the system. Zero-mean Gaussian noise with standard deviation ranging from 0 to 2 is added to the data, and SyMPLER is trained on two oscillation periods for different noise levels and ridge regularization strengths. The approximation accuracy is measured as the Euclidean distance between the concatenated vector of local model coefficients and the corresponding Taylor coefficient vector for the same approximation point, averaged over 30 repetitions to account for variability.

Figure~\ref{fig:pendulum_noisy_parms_distance} shows the mean coefficient distance as a function of noise level and regularization. For weak regularization, increasing noise leads to a progressive departure from the Taylor approximation due to sensitivity to outliers. Stronger regularization attenuates this effect by limiting parameter growth when fitting noisy samples, allowing the model to achieve an approximation error comparable to that of weakly regularized models under substantially lower noise intensities. However, strong regularization overly constrains the model at low noise levels, thus degrading the approximation in these scenarios. This highlights the importance of tuning the regularization parameter in SyMPLER.

\begin{figure}
    \centering
        \includegraphics[width=0.835\linewidth]{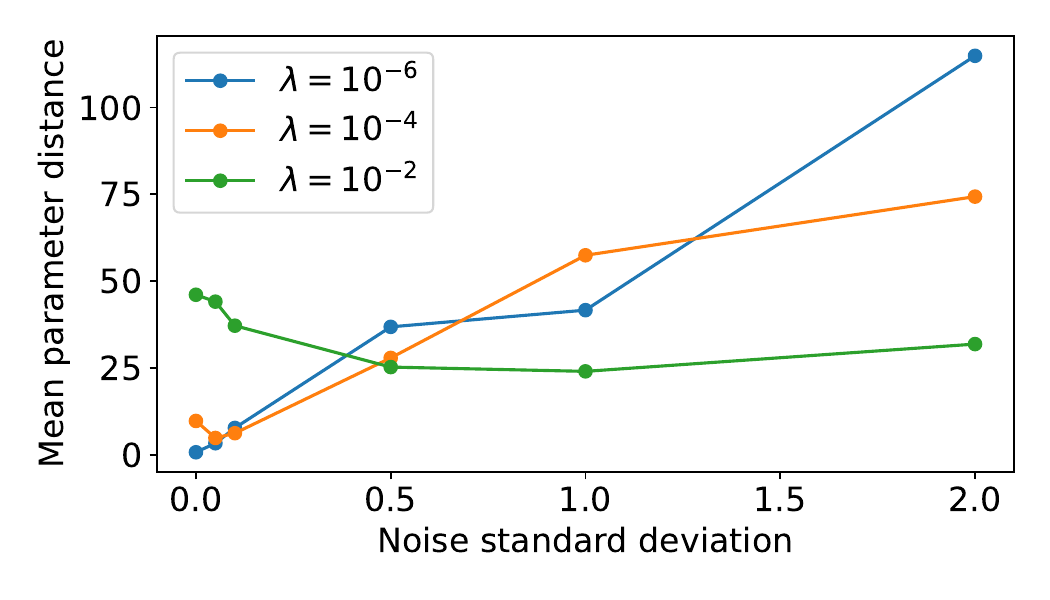}
       \caption{\rev{Mean distance between the vector of SyMPLER's coefficients and the vector of Taylor's coefficients for different levels of noise and regularization.}}
    \label{fig:pendulum_noisy_parms_distance}
\end{figure}

\end{revblock}

\section{Conclusion}
\label{sec:conclusion}


We introduced SyMPLER, a fully explainable continual learning model designed for time-series forecasting in nonstationary environments. By leveraging piecewise-linear approximations, SyMPLER dynamically adapts to new domains without requiring explicit clustering or user-defined thresholds. Experimental results demonstrate that SyMPLER outperforms evolving Takagi-Sugeno (eTS), Incremental Controller Networks (ICN) and EWC-SWFT models in electrical load forecasting, achieving accuracy comparable to state-of-the-art XGBoost trained offline while significantly reducing catastrophic forgetting. Additionally, SyMPLER provides human-interpretable insights into local dependencies between input features and predictions, bridging the gap between accuracy and explainability. Although the method currently focuses on handling covariate drifts, where the input distribution changes over time, its flexible structure makes it readily adaptable to address shifts in the functional relationship between inputs and outputs by modifying the local model selection criteria. Overall, SyMPLER presents a robust and interpretable solution for continual learning in time-series forecasting, offering a promising alternative to black-box models in real-world applications.
\bibliographystyle{IEEEtran}
\bibliography{bibtex/bib/bibliography}

\vspace{-0.25in}

\begin{IEEEbiography}[%
{\includegraphics[width=1in,height=1.25in,clip,keepaspectratio]{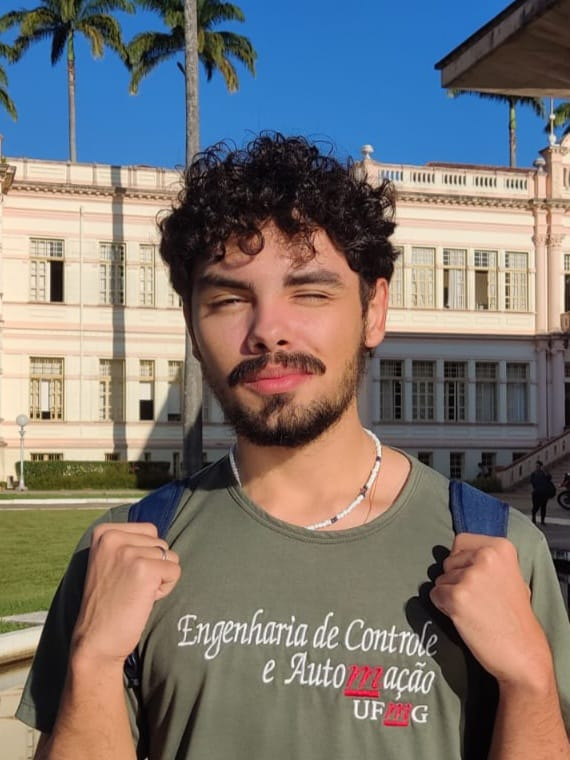}}%
]{Yan V. G. Ferreira} was born in Coronel Fabriciano, Brazil, in 2003. He is currently pursuing the B.S. degree in Control and Automation Engineering at the Federal University of Minas Gerais, Belo Horizonte, Brazil. His research interests include statistical learning theory, self-certified learning, continual learning, and interpretable machine learning.
\end{IEEEbiography}

\begin{IEEEbiography}[%
{\includegraphics[width=1in,height=1.25in,clip,keepaspectratio]{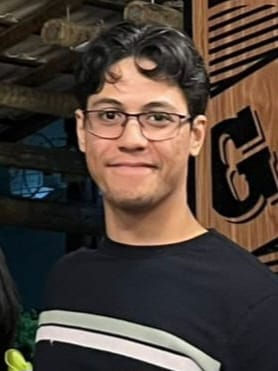}}%
]{Igor B. Lima} was born in Governador Valadares, Brazil, in 2001. He is currently pursuing the B.S. degree in Systems Engineering at the Federal University of Minas Gerais, Belo Horizonte, Brazil. His research interests include statistical learning theory, interpretable machine learning and information theory.
\end{IEEEbiography}

\begin{IEEEbiography}[%
{\includegraphics[width=1in,height=1.25in,clip,keepaspectratio]{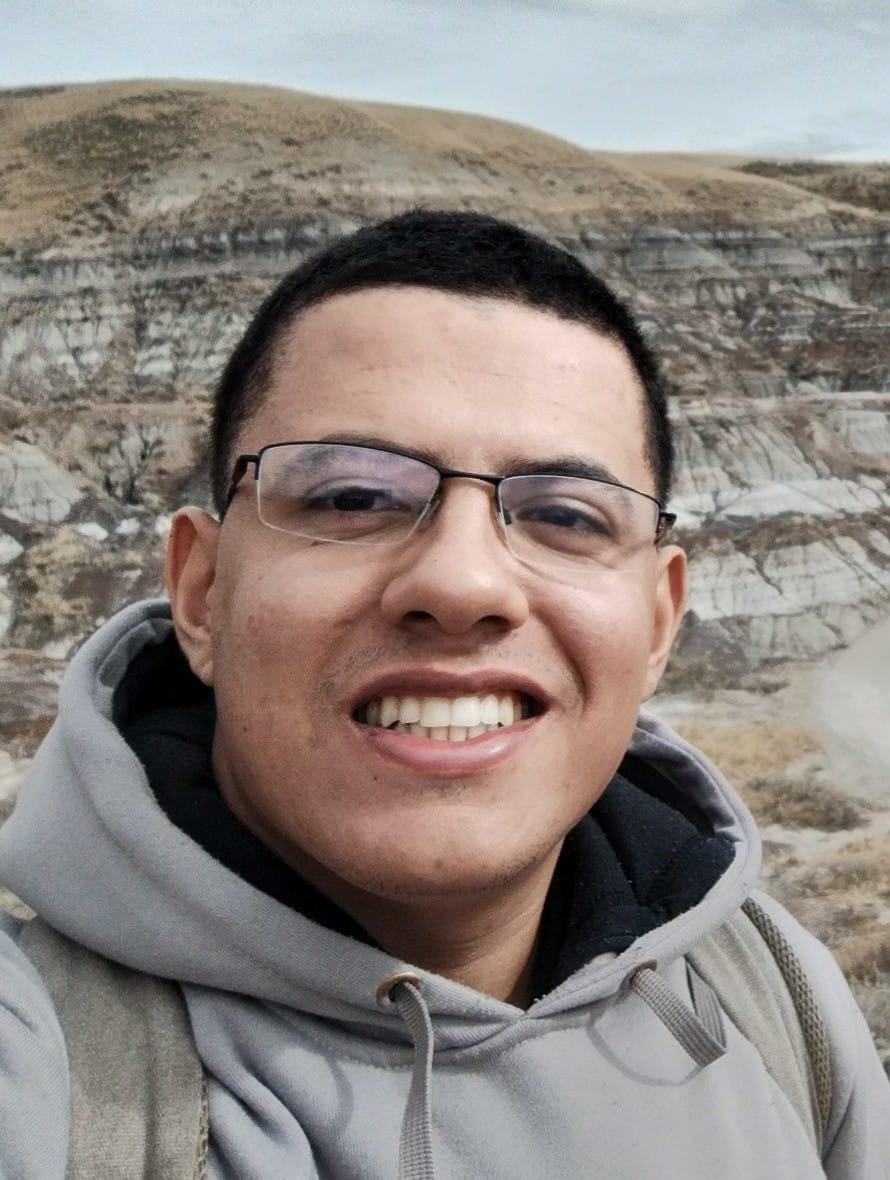}}%
] {Pedro H. G. Mapa S.} obtained his bachelor's degree in electrical engineering at Universidade Federal de Minas Gerais, and is a computing science graduate student at University of Alberta. His research interests are reinforcement leaning, continual learning, deep learning, and their combinations.
\end{IEEEbiography}

\begin{IEEEbiography}[%
{\includegraphics[width=1in,height=1.25in,clip,keepaspectratio]{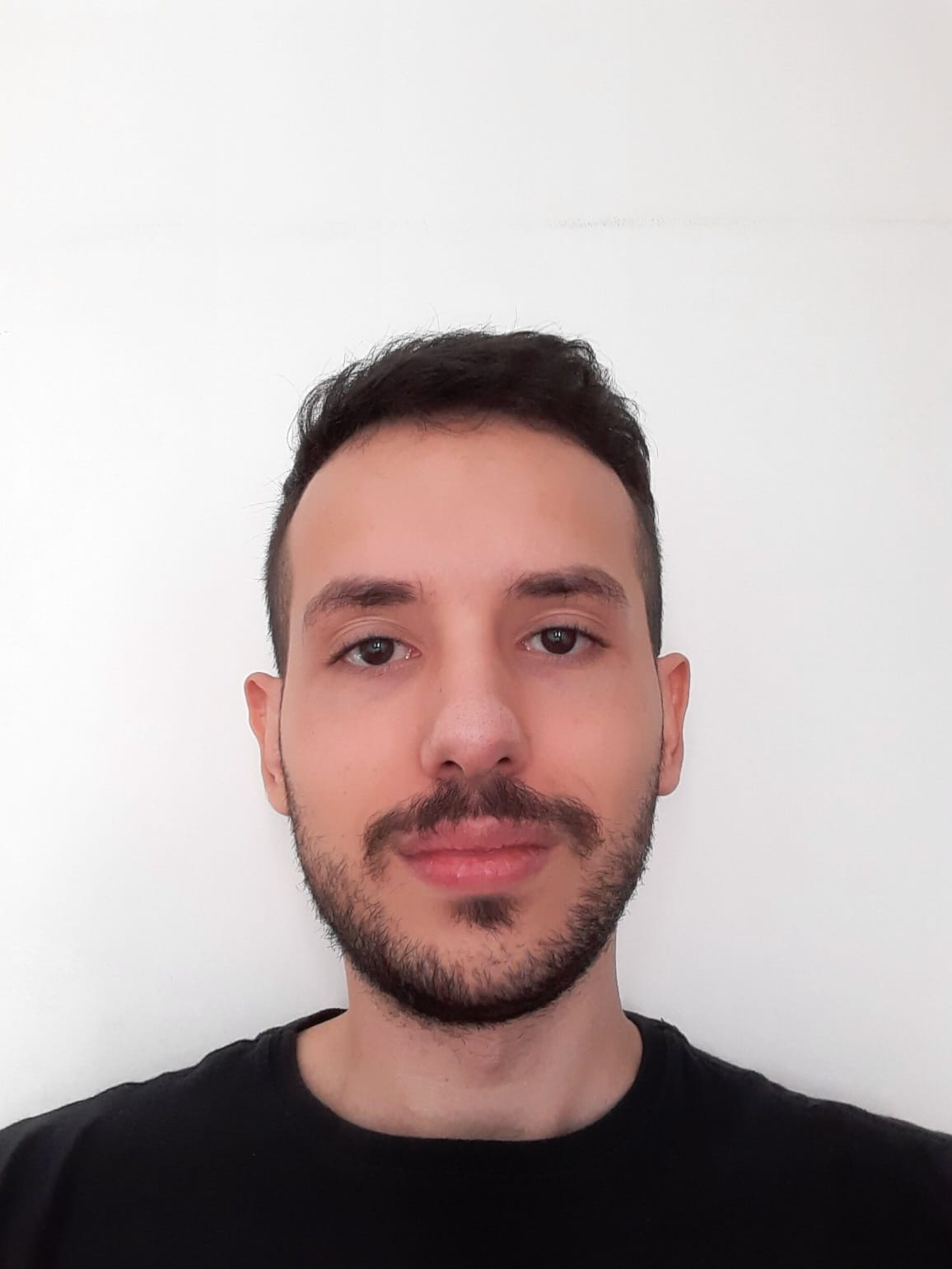}}%
] {Felipe V. Campos} is a master's student at the Federal University of Minas Gerais, Belo Horizonte, Brazil. His research focuses on machine learning and time series forecasting, with particular emphasis on Gaussian Process regression, signal processing, and automatic kernel composition methods. His current work explores heuristic-driven approaches combining wavelet decomposition and spectral analysis to improve the efficiency of kernel selection in probabilistic forecasting models. His broader research interests include optimization, mathematical modeling, and the intersection of statistical learning with signal processing techniques.
\end{IEEEbiography}

\begin{IEEEbiography}[%
{\includegraphics[width=1in,height=1.25in,clip,keepaspectratio]{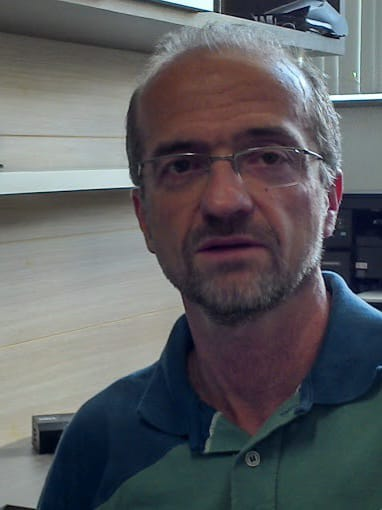}}%
]{Antonio P. Braga} received his degree in Electrical Engineering and his Master’s degree in Computer Science from Universidade Federal de Minas Gerais (UFMG), Brazil, in 1987 and 1991, respectively. His Ph.D. degree in Electrical Engineering was obtained from the University of London, Imperial College, in 1995, for his work in the Storage Capacity of Recurrent Neural Networks. Since 1991, he has been with the Electronics Engineering Department, Universidade Federal de Minas Gerais (UFMG), where he is a Full Professor and Head of the Computational Intelligence Laboratory. He is also an Associate Researcher of the Brazilian National Research Council. As a Professor and Researcher he has co-authored many books, book-chapters, journal and conference papers and has supervised many students and postdocs. He has served as Associate Editor of international journals such as IEEE Transactions on Neural Networks and Learning Systems, Pattern Recognition, Neural Processing Letters and Engineering Applications of Artificial Intelligence.
\end{IEEEbiography}

\end{document}